\documentclass{article}

\usepackage[english]{babel}

\usepackage[letterpaper,top=2cm,bottom=2cm,left=3cm,right=3cm,marginparwidth=1.75cm]{geometry}

\usepackage{graphicx}
\usepackage{authblk}
\usepackage[colorlinks=true, allcolors=blue]{hyperref}
\usepackage{amsmath}

\title{Search Engine and Recommendation System for the Music Industry built with JinaAI}

\author{
  {Ishita Gopalakrishnan}
  
    \textit \ \href{mailto:ishitagops12@gmail.com}{(ishitagops12@gmail.com)}
  \and
  {Sanjjushri Varshini R}
  
    \textit \ \href{mailto:sanjjushrivarshini@gmail.com}{(sanjjushrivarshini@gmail.com)}
  \and
  {Ponshriharini V}
  
    \textit \ \href{mailto:ponshriharini@gmail.com}{(ponshriharini@gmail.com)}
}

\begin{document}
\maketitle
\begin{abstract}
One of the most intriguing debates regarding a novel task is the development of search engines and recommendation-based systems in the music industry. Studies have shown a drastic depression in the search engine fields, due to concerning factors such as speed, accuracy and the format of data given for querying. Often people face difficulty in searching for a song solely based on the title, hence a solution is proposed to complete a search analysis through a single query input and is matched with the lyrics of the songs present in the database. Hence it is essential to incorporate cutting-edge technology tools for developing a user-friendly search engine. Jina AI is an MLOps framework for building neural search engines that are utilized, in order for the user to obtain accurate results. Jina AI effectively helps to maintain and enhance the quality of performance for the search engine for the query given. An effective search engine and a recommendation system for the music industry, built with JinaAI.
\end{abstract}

% \begin{keywords}


% \end{keywords}

\section{Introduction}

Developing an automated search engine for query analysis faces a lot of challenges in the real world especially with relevance to the music industry. Popularity bias is considered a huge downfall where query analysis always highlights a particular genre of music/musicians disregarding the mass public’s contributions. Moreover this hinders the fairness evaluation process for classifying song lyrics and other criterias pertaining to the music industry or any industry/sector in the real world. 

There is also a lack of importance and emphasis given to other regional and global languages when it comes to query analysis that has to be conquered to reach a larger audience and also improving diversity.
All musicians repeatedly examine and get to conclusions with their music scores. Hence it is vital to decide on user necessities, plan choices and software architecture of the search engine above metadata that exists. 

This system offers a solution to these problems by providing a search engine mechanism which searches for a song without any bias. It provides an impartial result which serves as a guiding medium to find songs of different underrated artists. This search engine works based on the song name, the song lyrics as well as the artists’ name. Different plots are drafted which are used to understand the data and its diversity. This system thus provides a bias-free search result which can be used to find songs in an objective manner.

\section{Literature Review}

% \subsection{How to create Sections and Subsections}

Getting a song’s lyrics from any given query word is the typical approach in the music Information retrieval system. But the problems which will be faced while trying to get the lyrics should also be taken into account. \cite{1} One such problem is the popularity bias. Popularity bias is something that should be taken into account since the recommendation system might recommend songs from famous and popular artists a lot more than the other not-so-popular artists which might be disadvantageous to similar, meritorious artists. It is found that the most accurate model (SLIM) produces the most popularity bias. \cite{2} Fairness in the music system recommendation is to be considered as well since the data is limited and it usually comes from the same source or is proprietary. This affects the users, item providers and the platform itself. Therefore, an approach is to be built to create a fair music recommendation system. 

The use of language models for generating lyrics and poetry has received increased interest in the last few years. \cite{9} A unique challenge relative to standard natural language problems, as their ultimate purpose is creativity; notions of accuracy and reproducibility are secondary to notions of lyricism, structure, and diversity. In this creative setting, traditional quantitative measures for natural language problems, such as BLEU (Bilingual Evaluation Understudy, which is a metric for automatically evaluating machine-translated text) scores, prove inadequate: a high-scoring model may either fail to produce output respecting the desired structure (e.g. song verses), be a terribly boring creative companion, or both. This paper proposes a mechanism for combining two separately trained language models into a framework that is able to produce output respecting the desired song structure while providing a richness and diversity of vocabulary that renders it more creatively appealing. Automation of text infilling enhances the user's inability as well as eases the laborious process. \cite{10} A simple approach for text infilling, the task of predicting missing spans of text at any position in a document. The authors aim to extend the capabilities of language models (LMs) to the more general task of infilling. Training off-the-shelf LMs on sequences containing the concatenation of artificially-masked text and the text which was masked provides an assistance writing tool. Infilling by language modelling can enable LMs to infill entire sentences effectively on three different domains: lyrics, short stories and scientific abstracts. Automated infilling of words is thoroughly resourceful. A beautiful emulsion between music and morality helps in a detailed analysis of moral values a person inhibits. \cite{11} An intricate union between music preferences and moral values by applying text analysis techniques to lyrics. The authors incorporated a machine learning framework designed to exploit regression approaches and evaluate the predictive power of lyrical features for inferring moral values. Music branches out to several different genres, enabling a broader research aspect of venturing into the Natural Language Processing field.
\cite{12} A novel approach for automatically classifying musical genres in Brazilian music using only the song lyrics. This kind of classification remains a challenge in the field of Natural Language Processing. To address this classification task, SVM, Random Forest and a Bidirectional Long Short-Term Memory (BLSTM) network was incorporated and combined with different word embedding techniques. Optimization of downstream tasks; supervised-learning tasks that utilize a pre-trained model or component, proves to be valuable. \cite{13} A novel task, Chorus Recognition, potentially benefits downstream tasks such as song search and music summarization. Different from the existing tasks such as music summarization or lyrics summarization relying on single-modal information, this models chorus recognition as a multi-modal one by utilizing both the lyrics and the tune information of songs. Incorporation of multi-modal Chorus Recognition model that considers diverse features. The empirical study performed on the dataset demonstrates that the approach outperforms several baselines in chorus recognition. In addition, improved the accuracy of its downstream task, song search by more than 10.6 percent.

\cite{7} The search engine performs well and we use tags to find them. Deep Learning techniques are used to automatically tag the important words. Their objective was to assess the impact of autonomously produced music tags on the musical content rankings provided by searches. The metrics used for evaluation purposes was Mean Average Precision before and after adding the tags. \cite{5} Using components like user modelling which involves user profile modelling and user listening experience modelling, and Item profiling which involves Editorial metadata, Cultural metadata, acoustic metadata and query type like entering the keyword or humming a song, it becomes more efficient to build a Music Information retrieval system. \cite{8} The tool Tambr translates writings into sound with chosen voices related to meaning and sentiment. Machine Learning based synthesizer search engine implementation of sentiment analysis is performed for producing the best results even which is not explored previously. The best synthesizers are picked whose definition agreeably deliberates the theories of the novel.\cite{4} Flow moods created a song recommendation system that provides songs based on users’ current moods. This is done by users selecting their current mood, annotating songs by experts, mood classification and playlist generation. \cite{6} All musicians repeatedly examine and get to conclusions with their music scores. Hence it is vital to decide on user necessities, plan choices and software architecture of the search engine above metadata that exists. The search engine responds to the search which is done by a collection of scores provided in MusicXML format. Reports are generated which will be beneficial for future enhancement.\cite{3} Their aim was to construct an Information Retrieval System from an android-based online music course application to explore and locate data situated in that music application. The tool used to develop was Android Studio software. The information retrieval serves index keywords into records that are stored in the Firebase database and then performs the query function to display the information associated with the query. Emotions/moods-based analysis helps in classifying a particular song in a better way. \cite{14} A profound method is proposed to detect the emotion of a song based on its lyrical and audio features. Lyrical features are generated by the segmentation of lyrics during the process of data extraction. 

\section{Tech Stack}

\subsection{Selenium:}

Selenium is an automation tool which is used to scrape data from various websites automatically. It is used in the system to scrape song lyrics from various websites such as azlyrics.com, lyrics.com, genius.com, etc. A total of 28372 song lyrics from different genres, radiating different emotions are scraped to create a bias-free dataset. 

\subsection{Jina AI - torch:}

Jina is an MLOps framework that is an open source tool which comprises advanced features built into it. It is a could-native neural search framework. It is used for building scalable deep learning search applications. It is the main tool that we have employed for building a search engine which will produce accurate results.
 
The following diagram depicts the flow diagram of the Jina AI working:

\begin{figure}[!htb]
\centering
\includegraphics[width=8cm,height=20cm,keepaspectratio]{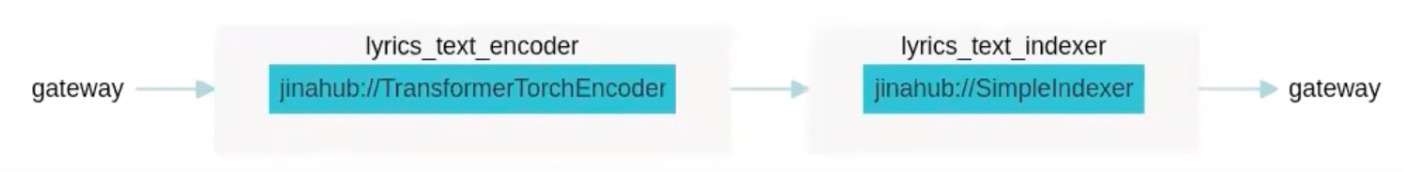}
\caption{\label{fig:links}Workflow of JinaAI}
\end{figure}

\subsection{Streamlit:}

Streamlit is an open-source application and it is used to build app frameworks for Machine Learning models.
Here, streamlit is used to display the visualization of data from the dataset. This lets us analyse the data and gather meaningful information from it. 

\subsection{Docker:}

Docker is an open-source tool used for the testing, development, shipping and running of the application. With the help of docker, the application was built faster. Since there exist multiple microservices the use of docker was mandatory. 

\subsection{ML Libraries:}

\subsubsection{Pandas:}

Pandas is a data manipulation tool which is built on top of the python programming language. It is fast, powerful, flexible and easy to use. It is used in the system to create data frames to manipulate the data present in the system. It makes data manipulation easy.

\subsubsection{Plotly:}

Plotly is a python library which is used for interactive data visualization purposes. It is used in the system to visualize the data present in the dataset in order to derive information and work accordingly.

\section{Methodology}

\begin{figure}[h]
\centering
\includegraphics[width=14cm,height=40cm,keepaspectratio]{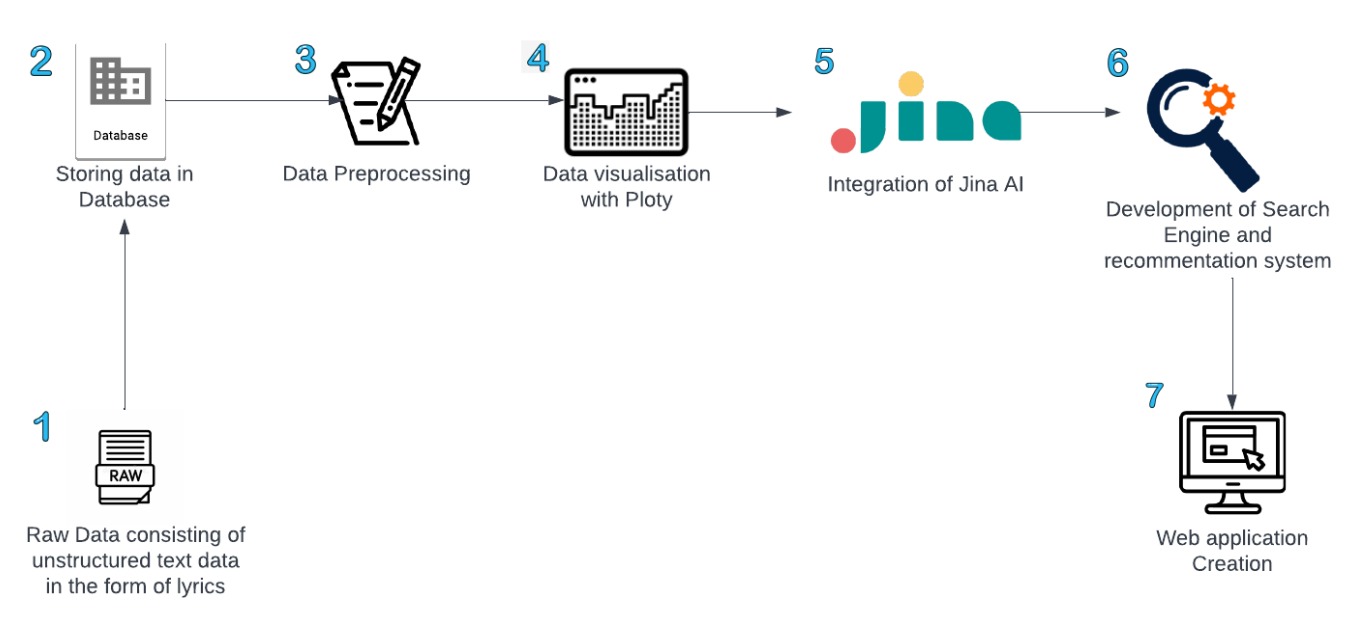}
\caption{\label{fig:links}Methodology Architecture}
\end{figure}

Working with song lyrics in real-time unstructured text data that has been processed and saved in a MongoDB database using Selenium for effective retrieval and management is involved in this. To ensure a clear and uniform format for subsequent analysis, the data processing procedures include noise removal, normalisation, tokenization, segmentation, lower casing, stop words removal, stemming, and lemmatization. To ensure unbiased datasets for the application, data visualisation is carried out using Plotly to produce graphs emphasising music genres and emotions. The application's central component, Jina AI, a deep learning-powered search engine, uses user searches to connect users with pertinent song lyrics and to build a recommendation system that suggests lyrics from various genres, emotions, and moods. The platform is transformed into an interactive, real-time web application with a simple dashboard that displays visualizations of top trending songs based on various criteria, enhancing user experience and interaction.

\subsection{Raw Data:}

The real-time unstructured text formatted data contains the lyrics of multiple songs.

\subsection{Database:}

Lyrics of songs are scraped using Selenium and stored in MongoDB for easier retrieval, and management of data.

\subsection{Data Processing:}

Data processing is one of the most essential processes in developing a machine learning model.
In order to feed information into a model for additional analysis and learning, text must first be transformed into a readable and consistent format.
The preprocessing phases include Noise removal, Normalization, Tokenization, Segmentation, Lower casing, Stop words removal, Stemming and Lemmatization.

\subsection{Data Visualization:}

The visualization of data is performed using Plotly. Various graphs highlighting the genre, emotions/ feelings are generated to deduce that there is no bias in the datasets used for the application.

\subsection{Integration of Jina AI:}

The whole application is based on Jina AI and is incorporated by matching a query word with the lyrics of a song. The search engine/recommendation-based system created helps in providing the best possible match for a word in the lyrics of a song.

\subsection{Development of search engine and recommendation system:}

A search engine using Jina AI is created to help match a query word with the lyrics of the song. A recommendation System is developed to generate and recommend lyrics of songs that can also contain different genres, emotions/ moods etc.

\subsection{Web application creation:}

The whole platform is converted to a real-time, dynamic web application that is user-friendly.
A dashboard is created to display the visualizations of the top trending songs based on genres, emotions, moods etc.

\section{Training Flow}

\begin{figure}[!htb]
\centering
\includegraphics[width=14cm,height=40cm,keepaspectratio]{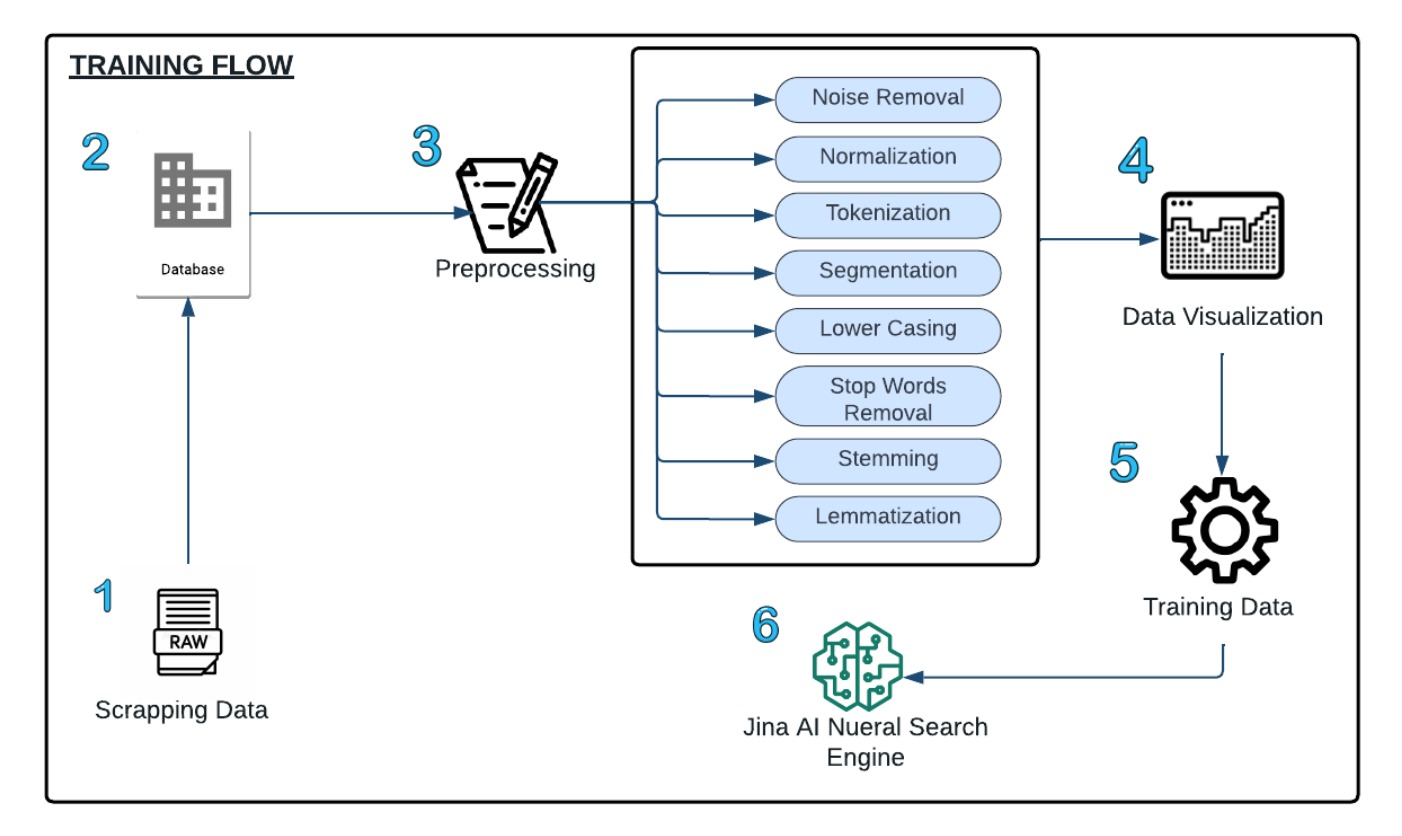}
\caption{\label{fig:links}Training Flow}
\end{figure}

\subsection{Scraping Data:}
The data is scraped using Selenium through different music-oriented websites that contain several lyrics of multiple songs.

\subsection{Database:}

The lyrics of the songs that are scraped using Selenium have then stored in a NoSQL, non-relational Database, MongoDB, where retrieval and managing of data become easier.

\subsection{Data/Text Preprocessing:}

In the context of natural language processing (NLP) and machine learning, effective data preprocessing techniques are crucial for refining raw text data to make it suitable for analysis. These techniques collectively enhance data quality and relevance, enabling various applications to yield better results. Noisy data, which contains incomprehensible or corrupted information, is a challenge for machines. This includes unstructured text devoid of coherent meaning. To address this, noisy segments are filtered out, leaving behind meaningful content. Normalization is a key step in data preparation. It involves transforming data values into a standardized range, often [0, 1], or onto a unit sphere. The goal is to ensure that different features of the data are on a similar scale. This enhances model performance and training stability. Tokenization involves breaking down paragraphs or sentences into smaller units called tokens. This is essential in NLP, allowing machines to process language by associating meaning with individual words. Tokenization is the initial phase, converting sentences into intelligible word units. Text segmentation divides lengthy documents into coherent and semantically meaningful segments. These segments aid in summarization, context comprehension, and other NLP applications. Effective segmentation enhances understanding of complex text data. Converting all text to lowercase simplifies subsequent analysis in NLP. By unifying letter cases, potential discrepancies are eliminated, enabling more accurate processing. Removing stop words, which have minimal semantic significance, prioritizes relevant content. This involves excluding tokens that match a predefined list of stop words. Removing non-informative words shifts focus to more meaningful terms.  Stemming reduces words to core roots by removing affixes. This simplifies text by unifying different forms of a word. Stemming aids in NLP tasks, helping search engines and chatbots understand word meanings. Lemmatization considers context to identify a word's base form. It accounts for the surrounding linguistic context to decipher word meanings. Both stemming and lemmatization aid in understanding language nuances in various applications. In summary, these data preprocessing techniques collectively refine raw text data for improved analysis. By eliminating noise, standardizing data, tokenizing text, segmenting content, adjusting cases, removing irrelevant words, and simplifying word forms, these techniques empower machines to extract valuable insights from text.

\subsection{Data Visualization:}

The data is henceforth converted into graphs using Plotly, for visualizing the results, produced based on genres, emotions, etc.

\subsection{Training Data:}

The data then undergoes training multiple times to maintain a good accuracy level.

\subsection{Jina AI Neural Search Engine:}

Jina AI neural search engine is trained and built for the data provided.

\section{Prediction Flow}

\begin{figure}[!htb]
\centering
\includegraphics[width=14cm,height=40cm,keepaspectratio]{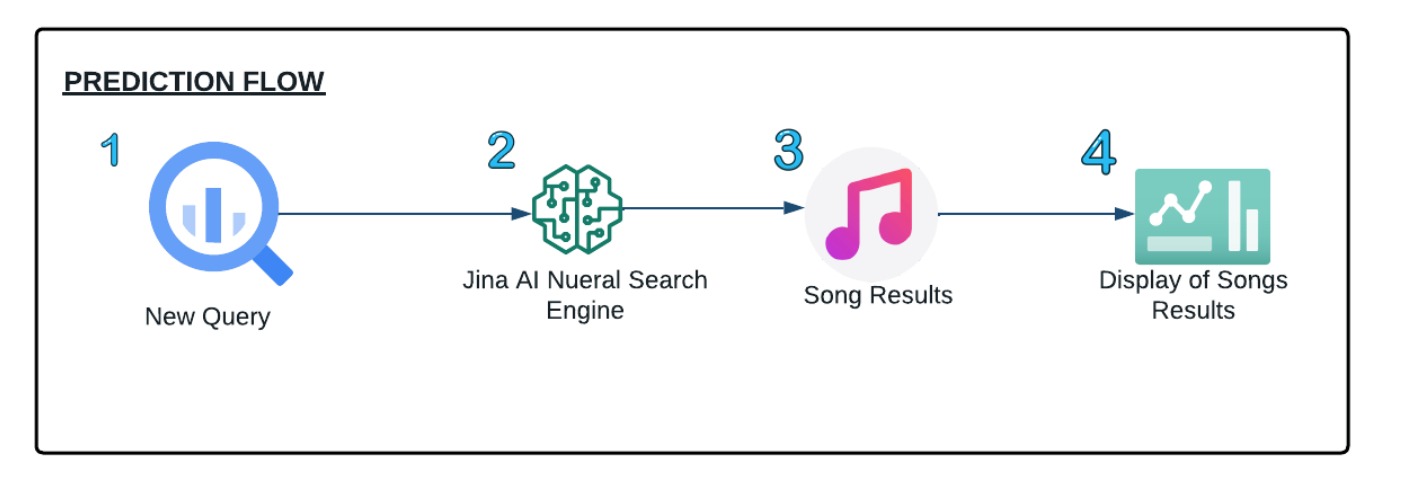}
\caption{\label{fig:links}Prediction Flow}
\end{figure}

\subsection{New Query: }

The query is in the form of a text/word that is given as input. The query word is then searched to match the word to all the lyrics of the song and display the results respectively.

\subsection{Jina AI Neural Search Engine:}

The Jina AI neural search engine created in the training phase is used for the new query generated by the user. This neural search engine works effectively for the query which has not been queried previously. 

\subsection{ Song Results and Display:}

The lyrics of the song that contains the query word within it are displayed to the user.

\section{Result Analysis}

The system first accepts any word as query input. This query input is then used to find songs. If the given query word is present in the song name or in the song lyrics itself, it displays the lyrics of that song. 
Thus, it accomplishes the job of a search engine. Here, the query word is ‘good’, hence it searches for all the songs that contain the word ‘good’ in it. The first result generated is the song ‘Good Life’ by Kehlani and G-Eazy which contains the word ‘good’ in its title as well as its lyrics.

\begin{figure}[!htb]
\centering
\includegraphics[width=14cm,height=40cm,keepaspectratio]{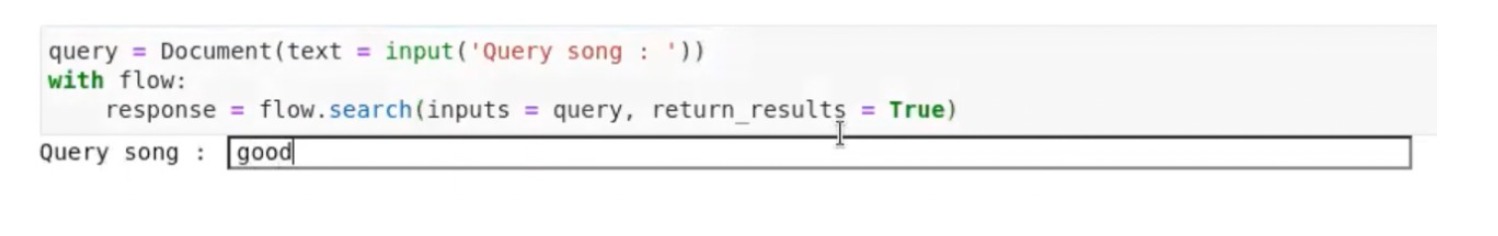}
\caption{\label{fig:links}Query Input of a particular song}
\end{figure}

The system operates as a search engine that takes any word as input and utilizes it to discover songs. By analyzing the input word, the system searches through its database for songs that either have the word in their title or within their lyrics. In essence, it functions as a music-centric search tool, seeking out songs that match the provided query. In this specific scenario, the query word is 'good.' Consequently, the system scans its collection of songs to locate all those containing the term 'good' either in their song titles or within the song lyrics. The first result generated by the search is the song 'Good Life' by Kehlani and G-Eazy, as it contains the word 'good' both in its title and its lyrical content.

In this scenario, the system serves as a search engine specifically designed for music-related queries. The user provides a word, in this case, the word 'good,' and the system's objective is to find songs that include this word either in their titles or within their lyrics.

To achieve this, the system has access to a vast database of songs with their corresponding metadata, including song titles and lyrics. When the user inputs the word 'good,' the system analyzes this query and starts searching its database for songs that match the provided word.

In the process of scanning the database, the system identifies all the songs that have the word 'good' either in their titles or in their lyrics. The first song that meets this criterion is 'Good Life' by Kehlani and G-Eazy. This song contains the word 'good' not only in its title but also within its lyrical content.

It's important to note that the search doesn't stop at 'Good Life.' There might be many other songs with the word 'good' in their titles or lyrics within the database. The given information only indicates the first result that the system presents in response to the query 'good.'

To summarize, the system is a music-centric search tool that uses the provided query to locate songs in its database containing the term 'good' either in their titles or lyrics. The first song it finds is 'Good Life' by Kehlani and G-Eazy, which contains 'good' both in its title and lyrics. The system can provide more results if the user wishes to explore other songs with the word 'good.'

\begin{figure}
    \centering
    \includegraphics[width=\linewidth,height=8cm]{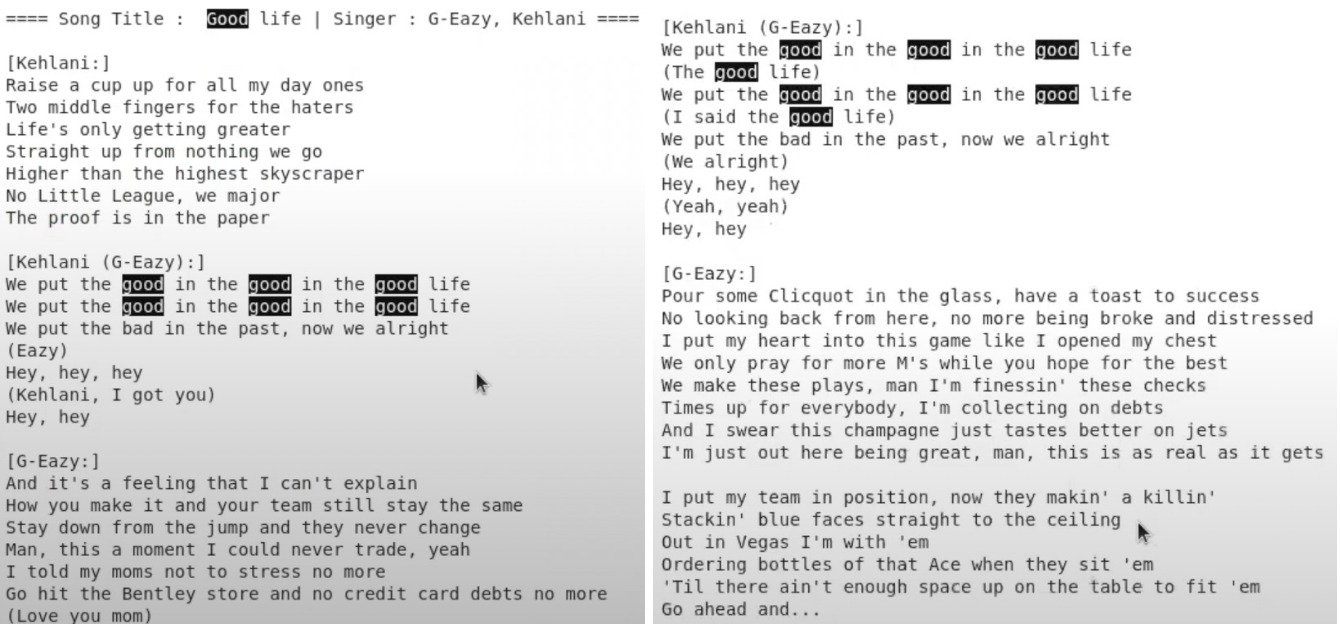}
    \caption{Lyric Query Detection}
    \label{fig:links}
 \end{figure}

In Figure 7, the dataset consists of song lyrics from the year 1950 to 2019. This provides a well-spread dataset which allows us to get results from over 60 years, thus making the search engine efficient enough to provide songs from every year. The year 2017 has the most songs amounting upto 660 songs from various different genres and emotions.

\begin{figure}[!htb]
\centering
\includegraphics[width=14cm,height=40cm,keepaspectratio]{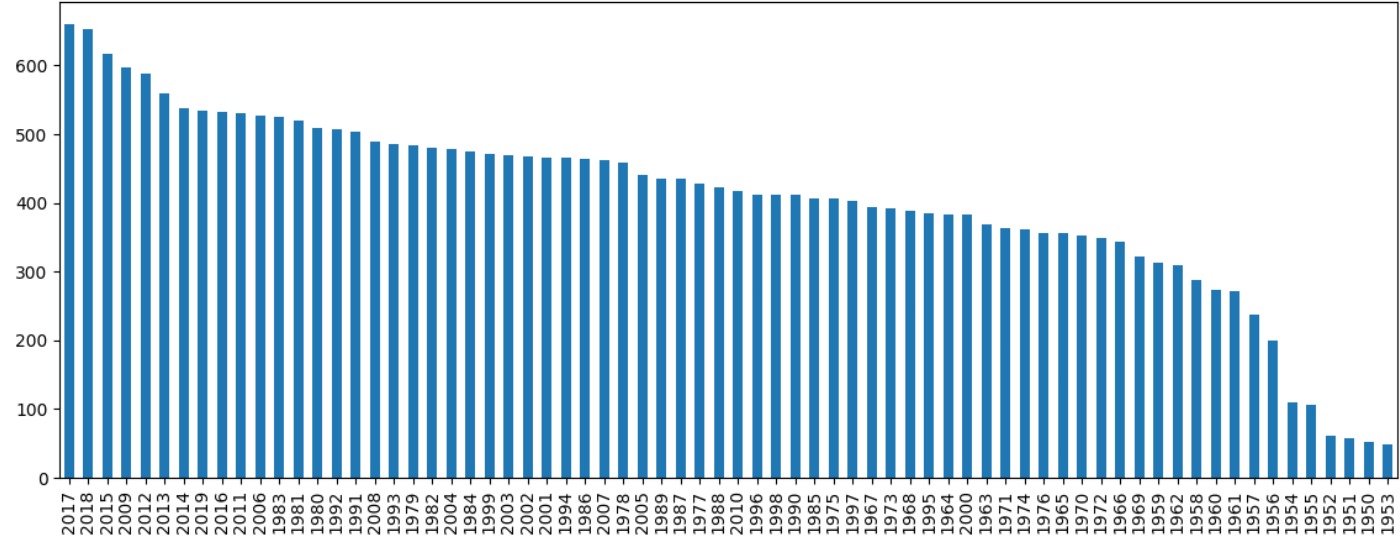}
\caption{\label{fig:links}Distribution of songs throughout the years}
\end{figure}

In Figure 8, the dataset has been categorized into 7 major genres namely - pop, country, blues, rock, jazz, reggae and hip hop. There are about 7042 songs under the genre pop. This includes pop songs from the years 1950 to 2019. We can infer that the dataset contains songs from various years and genres which means that the dataset is not biased.

\begin{figure}[!htb]
\centering
\includegraphics[width=14cm,height=40cm,keepaspectratio]{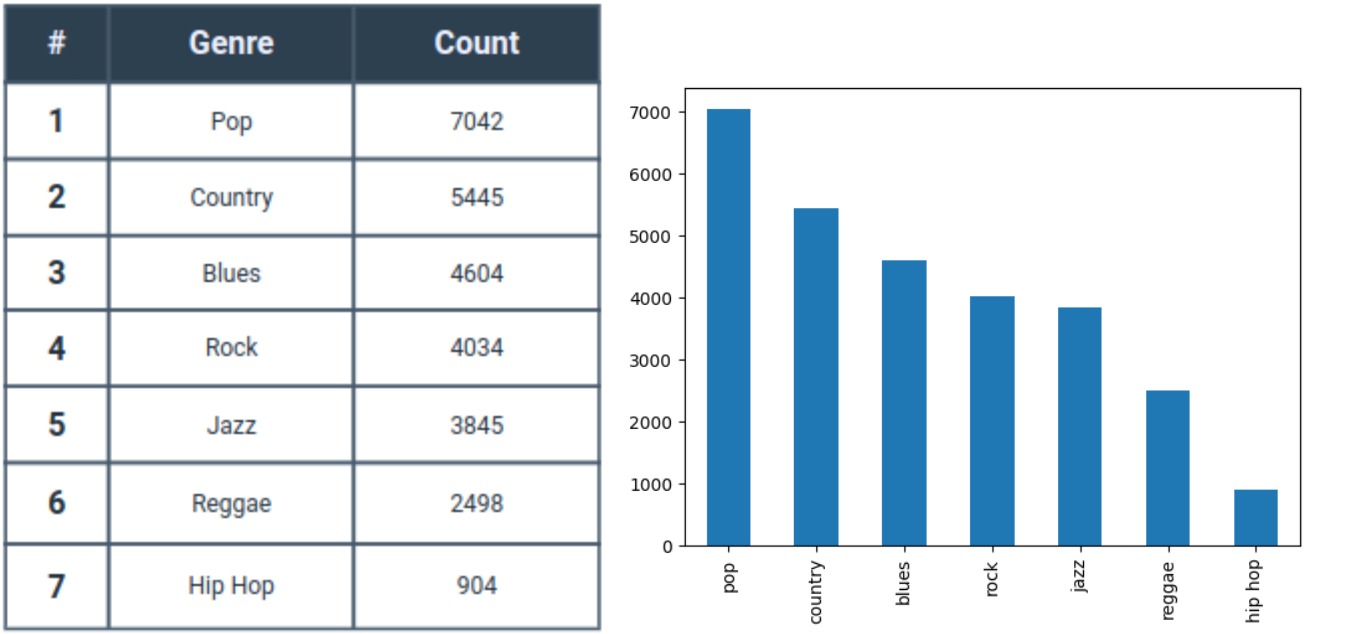}
\caption{\label{fig:links}Distribution of songs based on genres}
\end{figure}

In Figure 9, the dataset has been categorized into 8 significant emotions/ moods, namely - sadness, violence, world/ life, obscene, music, night/ time, romantic and feelings. There are about 6000 songs under the sadness emotion. We can deduce that the dataset contains themes harbouring emotions/ moods, concluding with the fact that the dataset is not biased.

\begin{figure}[!htb]
\centering
\includegraphics[width=14cm,height=40cm,keepaspectratio]{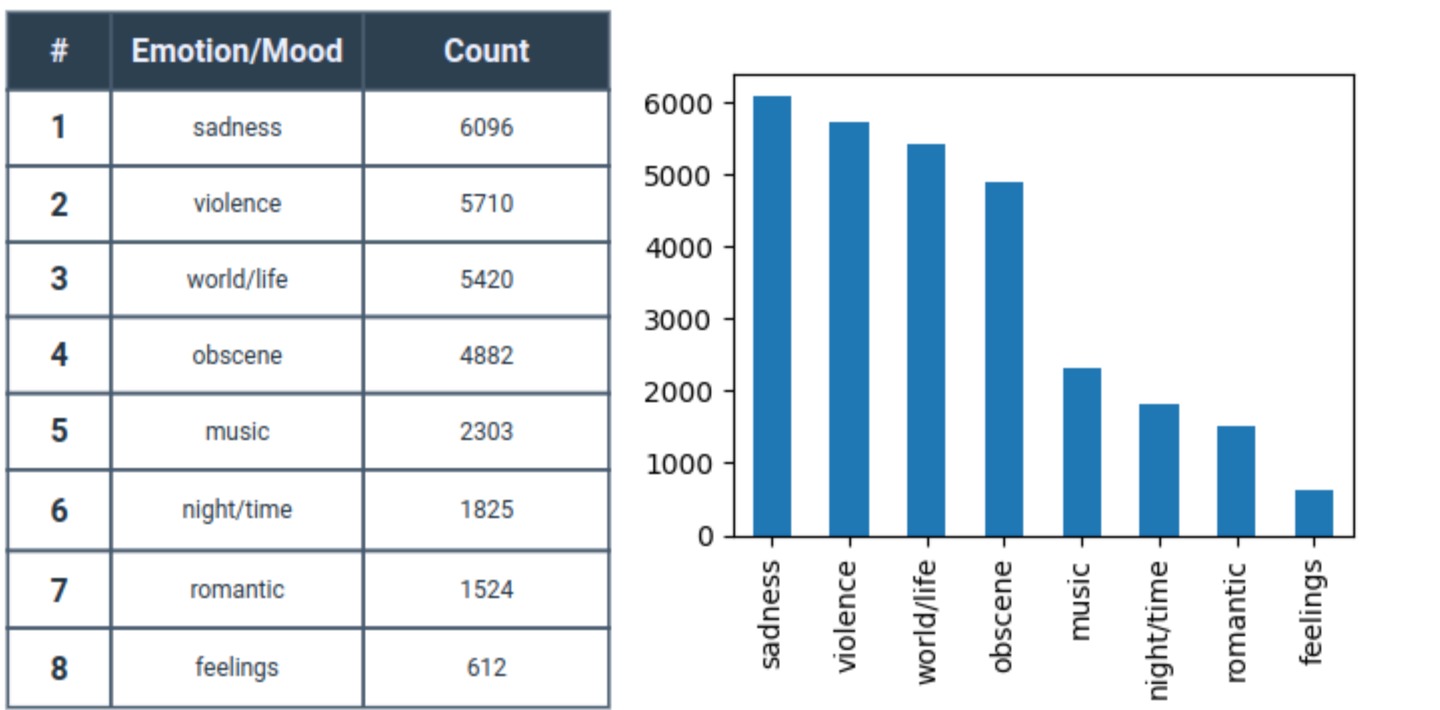}
\caption{\label{fig:links}Distribution of songs based on emotions/moods}
\end{figure}

\section{Future implementations}

There are several improvements and future implementations that can be proposed over the course of building a search engine with a recommendation-based system using Jina AI. One of the approaches involves generating a genre by searching the lyrics and categorizing them respectively. Building a system with emotion as one of the parameters will improve the model as searching for songs or suggesting songs based on emotions can also serve as a new feature. This also involves categorizing songs according to the mood they radiate. The top trending songs could be produced through Emotional and mood-based analysis. 
Songs recommended through uploaded images pave a new way to improve the application.

\section{Conclusion}

The main tenet of this applied research is to create a search engine which makes the process of searching for song lyrics a lot more efficient, fast and easy. One of the major pros of the system is to give a profound result for the query-based search analysis. JINA AI neural search model is a beneficial approach to the research as it is semantic and agnostic, and the use of a wide array of models makes the neural search so much easier to work on. JINA AI facilitates and works with dealing with infrastructure and optimization problems. It is designed to provide search systems for all types of data, including text, photos, audio, video, and a variety of others. You can use effective patterns to piece together the system using the modular design and multi-layer abstraction, or you can chain them together into a flow for an end-to-end experience. Thus, JINA gives us a fully customizable and novel approach.

% \subsection{How to add Comments and Track Changes}

% Comments can be added to your project by highlighting some text and clicking ``Add comment'' in the top right of the editor pane. To view existing comments, click on the Review menu in the toolbar above. To reply to a comment, click on the Reply button in the lower right corner of the comment. You can close the Review pane by clicking its name on the toolbar when you're done reviewing for the time being.

% Track changes are available on all our \href{https://www.overleaf.com/user/subscription/plans}{premium plans}, and can be toggled on or off using the option at the top of the Review pane. Track changes allow you to keep track of every change made to the document, along with the person making the change. 

% \subsection{How to add Lists}

% You can make lists with automatic numbering \dots

% \begin{enumerate}
% \item Like this,
% \item and like this.
% \end{enumerate}
% \dots or bullet points \dots
% \begin{itemize}
% \item Like this,
% \item and like this.
% \end{itemize}

% \subsection{How to add Citations and a References List}

% You can simply upload a \verb|.bib| file containing your BibTeX entries, created with a tool such as JabRef. You can then cite entries from it, like this: \cite{greenwade93}. Just remember to specify a bibliography style, as well as the filename of the \verb|.bib|. You can find a \href{https://www.overleaf.com/help/97-how-to-include-a-bibliography-using-bibtex}{video tutorial here} to learn more about BibTeX.

% If you have an \href{https://www.overleaf.com/user/subscription/plans}{upgraded account}, you can also import your Mendeley or Zotero library directly as a \verb|.bib| file, via the upload menu in the file-tree.


\begin{thebibliography}{00}

\bibitem{1} Douglas R. Turnbull, Sean McQuillan, Vera Crabtree, John Hunter, and Sunny Zhangl. "Exploring Popularity Bias in Music Recommendation Models and Commercial Steaming Services." arXiv preprint arXiv:2208.09517 (2022).

\bibitem{2} Dinnissen, Karlijn, and Christine Bauer. "A Stakeholder-Centered View on Fairness in Music Recommender Systems." arXiv preprint arXiv:2209.06126 (2022).

\bibitem{3} Andika, Albertus Dera, Farid Baskoro, and Eppy Yundra. "Application of Retrieval Information on Android-Based Online Music Course Application." International Joint Conference on Science and Engineering (IJCSE 2020). Atlantis Press, 2020.

\bibitem{4} Théo Bontempelli, Benjamin Chapus, François Rigaud, Mathieu Morlon, Marin Lorant, Guillaume Salha-Galvan. "Flow Moods: Recommending Music by Moods on Deezer." Proceedings of the 16th ACM Conference on Recommender Systems. 2022. 

\bibitem{5} Song, Yading, Simon Dixon, and Marcus Pearce. "A survey of music recommendation systems and future perspectives." 9th international symposium on computer music modeling and retrieval. Vol. 4. 2012.

\bibitem{6} Bahraini, Arman, and Eli Tilevich. "Ask toscanini! architecting a search engine for music scores beyond metadata." Proceedings of the 34th ACM/SIGAPP Symposium on Applied Computing. 2019.

\bibitem{7} Weberyd, Emma. "Evaluating the Impact of Automated Music Tags on Search Engine Ranking Quality." (2020).

\bibitem{8} Salas, Jessie. "Generating music from literature using topic extraction and sentiment analysis." IEEE Potentials 37.1 (2018): 15-18.

\bibitem{9} Castro, Pablo Samuel, and Maria Attarian. "Combining learned lyrical structures and vocabulary for improved lyric generation." arXiv preprint arXiv:1811.04651 (2018).

\bibitem{10} Donahue, Chris, Mina Lee, and Percy Liang. "Enabling language models to fill in the blanks." arXiv preprint arXiv:2005.05339 (2020).

\bibitem{11} Preniqi, Vjosa, Kyriaki Kalimeri, and Charalampos Saitis. "" More Than Words": Linking Music Preferences and Moral Values Through Lyrics." arXiv preprint arXiv:2209.01169 (2022).

\bibitem{12} Raul de Araújo Lima, Rômulo César Costa de Sousa, Simone Diniz Junqueira Barbosa, Hélio Cortês Vieira Lopes. "Brazilian lyrics-based music genre classification using a BLSTM network." Artificial Intelligence and Soft Computing: 19th International Conference, ICAISC 2020, Zakopane, Poland, October 12-14, 2020, Proceedings, Part I 19. Springer International Publishing, 2020.

\bibitem{13} Jiaan Wang, Zhixu Li, Binbin Gu, Tingyi Zhang, Qingsheng Liu, Zhigang Chen. "Multi-modal Chorus Recognition for Improving Song Search." Artificial Neural Networks and Machine Learning–ICANN 2021: 30th International Conference on Artificial Neural Networks, Bratislava, Slovakia, September 14–17, 2021, Proceedings, Part I 30. Springer International Publishing, 2021.

\bibitem{14} Adit Jamdar, Jessica Abraham, Karishma Khanna, Rahul Dubey. "Emotion analysis of songs based on lyrical and audio features." arXiv preprint arXiv:1506.05012 (2015).

\end{thebibliography}
\end{document}